\documentclass{article}

\usepackage{spconf}
\usepackage{amsmath}
\usepackage{graphicx}
\usepackage{amsfonts}
\usepackage{subcaption}
\usepackage{multirow}
\usepackage{xcolor}
\usepackage{enumitem}
\usepackage{hyperref}
\usepackage{makecell}
\usepackage{tabularx}
\usepackage{colortbl}
\usepackage{natbib}
\usepackage{algorithm}
\usepackage{algpseudocode}
\usepackage{pifont}
\usepackage{amssymb}
\usepackage{bbm}
\newcolumntype{x}[1]{>{\centering\let\newline\\\arraybackslash\hspace{0pt}}p{#1}}

\DeclareRobustCommand{\itshape}{%
  \not@math@alphabet\itshape\mathit
  \fontshape\itdefault\selectfont
  \color{gray}%
}

\definecolor{darkgreen}{rgb}{0.0, 0.4, 0.13}

\newcommand{\cmark}{\ding{51}}%
\newcommand{\xmark}{\ding{55}}%

\title{Investigating Self-Supervised Methods for Label-Efficient Learning}

\name{Srinivasa Nandam$^{a}$ \qquad Sara Atito$^{a,b}$ \qquad Zhenhua Feng$^{b}$ \qquad Josef Kittler$^{b}$ \qquad Muhammad Awais$^{a,b}$ \vspace{-0.3cm}}

\address{$^{a}$ Surrey Institute for People-Centred AI, University of Surrey, Guildford, GU2 7XH, UK\\
         $^{b}$ Centre for Vision, Speech and Signal Processing (CVSSP), University of Surrey \vspace{-0.2cm}}

\begin{document}
%
\maketitle

\begin{abstract}
Vision transformers combined with self-supervised learning have enabled the development of models which scale across large datasets for several downstream tasks like classification, segmentation and detection. The low-shot learning capability of these models, across several low-shot  downstream tasks, has been largely under explored. We perform a system level study of different self supervised pretext tasks, namely contrastive learning, clustering, and masked image modelling for their low-shot capabilities by comparing the pretrained models.
In addition we also study the effects of collapse avoidance methods, namely centring, ME-MAX, sinkhorn, on these downstream tasks. Based on our detailed analysis, we introduce a framework involving both mask image modelling and clustering as pretext tasks, which performs better across all low-shot downstream tasks, including multi-class classification, multi-label classification and semantic segmentation. Furthermore, when testing the model on full scale datasets, we show performance gains in multi-class classification, multi-label classification and semantic segmentation.


\end{abstract}


\section{Introduction}
\label{sec:intro}

Self-supervised Learning (SSL) has gained popularity as a learning technique for acquiring meaningful representations in an unsupervised manner
By training on large unlabeled datasets using self-supervised pretext and auxiliary tasks, SSL produces features that can efficiently be applied to downstream tasks with fewer labels, as demonstrated in \cite{sit, dino, msn, mae, simmim, mcssl}. Furthermore, SSL has enabled Vision Transformers (ViTs)~\cite{vit} to outperform Convolutional Neural Networks (CNNs) in various image-related tasks, including classification, detection, and segmentation~\cite{sit, simmim, mae, ibot}. 

\begin{figure}[h]
        \centering
        \includegraphics[width=\linewidth]{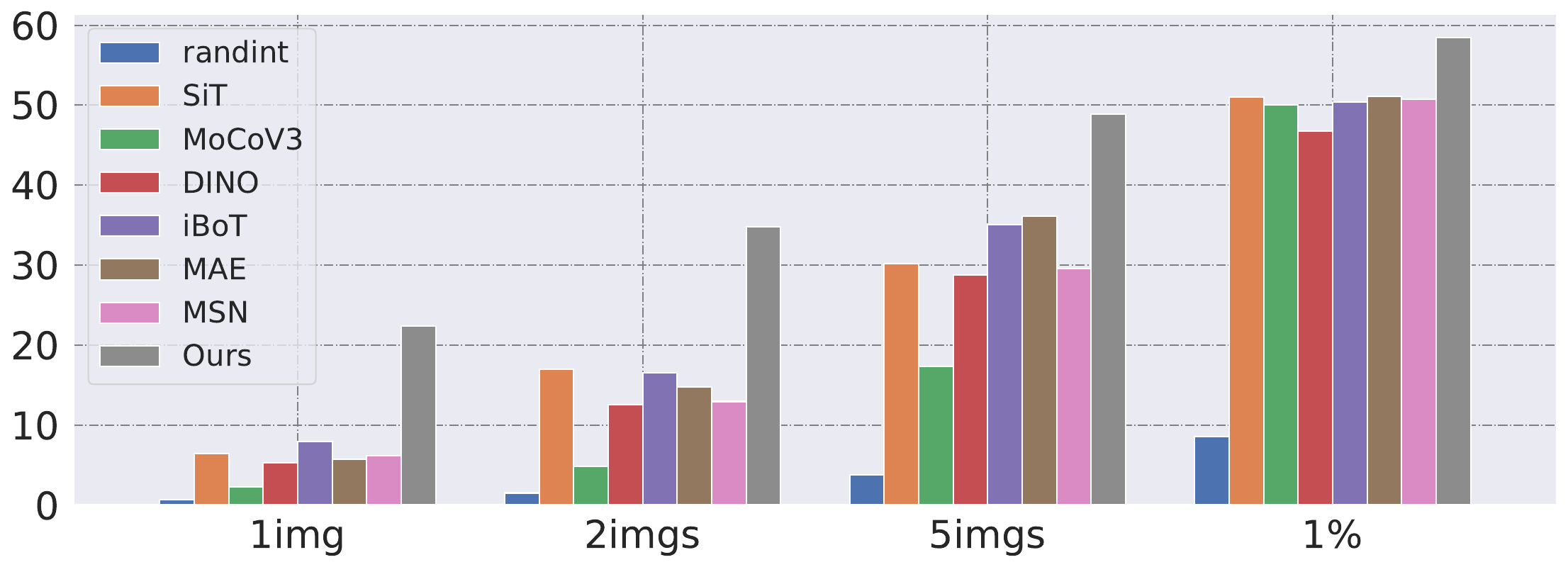} 
        \caption{Accuracy of different methods on low-shot classification task for 1, 2, 5 images, and 1\% of ImageNet-1K.}
        \label{fig:plot_low-shot}
        \vspace{-0.6cm}
\end{figure}

Self-supervised methods for modelling global discriminative features often employ either contrastive pretext tasks \cite{simclr, simclr2, moco, mocov2, mocov3} or clustering pretext tasks~\cite{dino, swav, msn}. On the other hand, an alternative avenue of SSL, i.e. Masked Image Modeling (MIM) methods, has emerged with a distinct focus on capturing contextual information by reconstruction either at the pixel level~\cite{gmml, simmim, mae} or the token level~\cite{maskfeat, beit}, thereby lacking the incorporation of discriminative details that are crucial for generating globally informative features. \citep{lg} explored a combination of contrastive pretext tasks and MIM in the pixel space, which has shown to yield improved representations. 
 iBoT~\cite{ibot} follows a similar approach, but replace the contrastive task with a clustering task at both global and patch levels and utilise MIM for token-level masked region prediction. Further, MSN~\cite{msn} proposes ME-MAX loss instead of the centring trick proposed in DINO \cite{dino} for better low-shot linear evaluation.


All previous methods have demonstrated strong performance in the large dataset size regime. However, their performance in low-shot scenarios has been largely overlooked, with the notable exception of the MSN approach. However, MSN do not analyse the effect of different SSL components, like the choice of pretext task, and collapse avoidance mechanisms for low-shot learning. Moreover, MSN confines its evaluation to low-shot linear classification on ImageNet-1K, which presents several limitations. For one, pre-training with a contrastive or invariant loss function typically results in high linear evaluation performance, especially when the model is assessed using the same dataset on which it was pre-trained. This outcome can falsely imply effectiveness in low-shot scenarios. Additionally, this narrow method of assessment does not accurately predict the model's transferability to different tasks and datasets. Hence, there's a notable gap in the literature in comprehensive, system-level analyses of the impact of SSL and its components on low-shot applications.

In this paper, we perform a detailed study of the impact of different pretext tasks and the choice of a collapse avoidance method on the performance of low-shot downstream tasks. In addition, we also study the effect of extending the instance discrimination pretext task to the patch level. We provide an overview of different pretext tasks and collapse avoidance mechanisms used by previous frameworks in Table~\ref{table:comp}.

Based on the above analysis, we investigate a simple model with a combination of two different pretext tasks namely clustering and MIM for low-shot learning. Clustering is done at both, the class token level to capture global semantics and the patch level to capture local semantics. We perform MIM at pixel level, in addition to clustering, to capture finegrained details.  When evaluated on several low-shot downstream tasks namely multi-label classification, multi-class classification and semantic segmentation, the proposed simple model works better due its ability to capture details at various levels. We also present the performance of state-of-the-art self-supervised models on these downstream tasks. Figure~\ref{fig:plot_low-shot} shows the performance comparison of various SSL methods in low-shot classification on ImageNet-$1$K. To analyse the scaling behaviour on full datasets we finetune the model on standard finetuning evaluation settings following the previous SSL approaches \cite{dino, mcssl, ibot}. We find that our model performs favourably in these settings as well.

\begin{table}[h!]
        \centering
        \resizebox{\linewidth}{!} {
        \begin{tabular}{l | c c c | c c c } 
        \hline
        Method & \cellcolor{gray!40} ME-MAX & \cellcolor{gray!40} Sinkhorn & \cellcolor{gray!40} Centring & Contrastive & Clustering & MIM\\  
        \hline
        MoCoV3~\cite{mocov3} & \xmark & \xmark & \xmark & cls & \xmark & \xmark \\
        DINO~\cite{dino} & \xmark & \xmark & \cmark & \xmark & cls & \xmark \\
        iBoT~\cite{ibot} & \xmark & \xmark & \cmark & \xmark & cls+patch & \xmark \\
        MSN~\cite{msn} & \cmark & \cmark & \xmark & \xmark & cls & \xmark \\
        MAE~\cite{mae} & \xmark & \xmark & \xmark & \xmark & \xmark & \cmark \\
        \rowcolor{blue!20} Ours & \cmark & \xmark & \xmark & \xmark & cls+patch & \cmark \\
        \hline
    \end{tabular}
    }
    \caption{A review of different self supervised methods and their pretext tasks and  collapse avoidance mechanisms.}
    \label{table:comp}
    \vspace{-0.5cm}
\end{table}

\section{Related Works}
\label{sec:related_works}

The early SSL methods in computer vision relied on simple pretext tasks, such as solving jigsaw puzzles \cite{jigsaw}, predicting colour from grayscale images \cite{unsup1}, or classifying relative positions \cite{relpos}. However, recent advances introduced more sophisticated pretext tasks with complex training objectives. Generative methods, that mask parts of the input randomly and predict those regions at the pixel or token level, gained popularity in SSL \cite{sit, dae, igpt, gmml, simmim, mae}. GMML in SiT~\cite{sit} was the first ViT method to demonstrate that masked autoencoder, i.e., masking randomly large proportions of image patches and reconstructing them, leads to strong self-supervised pretext task capable of outperforming supervised pretraining. BeiT~\cite{beit} extended the idea of masked autoencoder using a discrete variational autoencoder (dVAE) for token generation and prediction. SimMIM~\cite{simmim} and MAE~\cite{mae} employed the idea of heavy masking and recovery of information to a larger scale using an autoencoder-style approach for pixel-wise reconstruction. GMML, MAE, and SimMIM reconstruct at pixel level, 
whereas BeiT reconstructs at token level.
These methods do not enforce global level representation consistency across different views of the same image.

Contrastive methods, on the other hand, learn invariance by emphasising similarity between positive views and reducing similarity between negative views using InfoNCE loss~\cite{infonce}. SimCLR~\cite{simclr} highlighted the importance of data augmentation, while MoCo~\cite{moco} introduced a memory bank to address the 
issue of large batch size. SiT~\cite{sit} combined MIM with contrastive learning, leading to performance improvements. Clustering-based methods achieved invariance by learning similar cluster assignments for different augmented views. SwAV~\cite{swav} used cluster assignments as a supervisory signal, and DINO~\cite{dino} emphasised the role of momentum encoder and multiple crops for SSL.
Collapse is a major issue for self supervised methods where trivial solutions are produced by the network where embeddings do not have enough variance in the representation space. Recent methods have used asymmetry in design~\cite{simsiam}, sinkhorn to normalise the teacher cluster assignments~\cite{swav}, momentum encoders to generate target embeddings~\cite{moco}, centring to make teacher distribution more uniform along with sharpening~\cite{dino} to avoid collapse. 
MSN \cite{msn} is a variant of DINO, where collapse is avoided with ME-MAX 
loss instead of centring or sinkhorn, showing superior performance in low-shot linear evaluation. By default they apply ME-MAX with sinkhorn to avoid setting the scaling factor for ME-MAX loss. 
iBoT \cite{ibot} extends DINO with masking and clustering applied to both patch and class tokens, yet iBoT lacks fine-grained context due to lack of pixel reconstruction.

Semi-supervised methods~\cite{semisup, semisup1} have considered extreme low data scenarios which have been mostly overlooked by SSL community with the exception of MSN~\cite{msn}. Moreover, the existing literature lacks a comprehensive examination of the impact of various SSL components on low-shot learning performance. While MSN demonstrates superior low-shot linear evaluation through ME-MAX loss utilization, it falls short in exploring the influence of diverse pretext tasks and collapse avoidance mechanisms on various low-shot downstream tasks. Additionally, their exclusive focus on low-shot linear evaluation using the pretraining dataset (ImageNet-1K) may not generalize effectively to different datasets and downstream tasks. Hence, our emphasis is on low-shot finetuning across diverse tasks and datasets. 
We perform thorough analysis of different components including the choice of pretext task, choice of collapse avoidance. Motivated by our findings we propose a method which has multiple pretext tasks: clustering and masked image modelling. The introduced model applies clustering on both class and patch tokens and does reconstruction with a pixel level loss. When compared to other SSL methods we find that our model performs the best across several low-shot downstream tasks. The performance also scales to large scale when performing finetuning on full datasets. 


\begin{figure*}[h]
\centering
\includegraphics[width=0.8\linewidth]{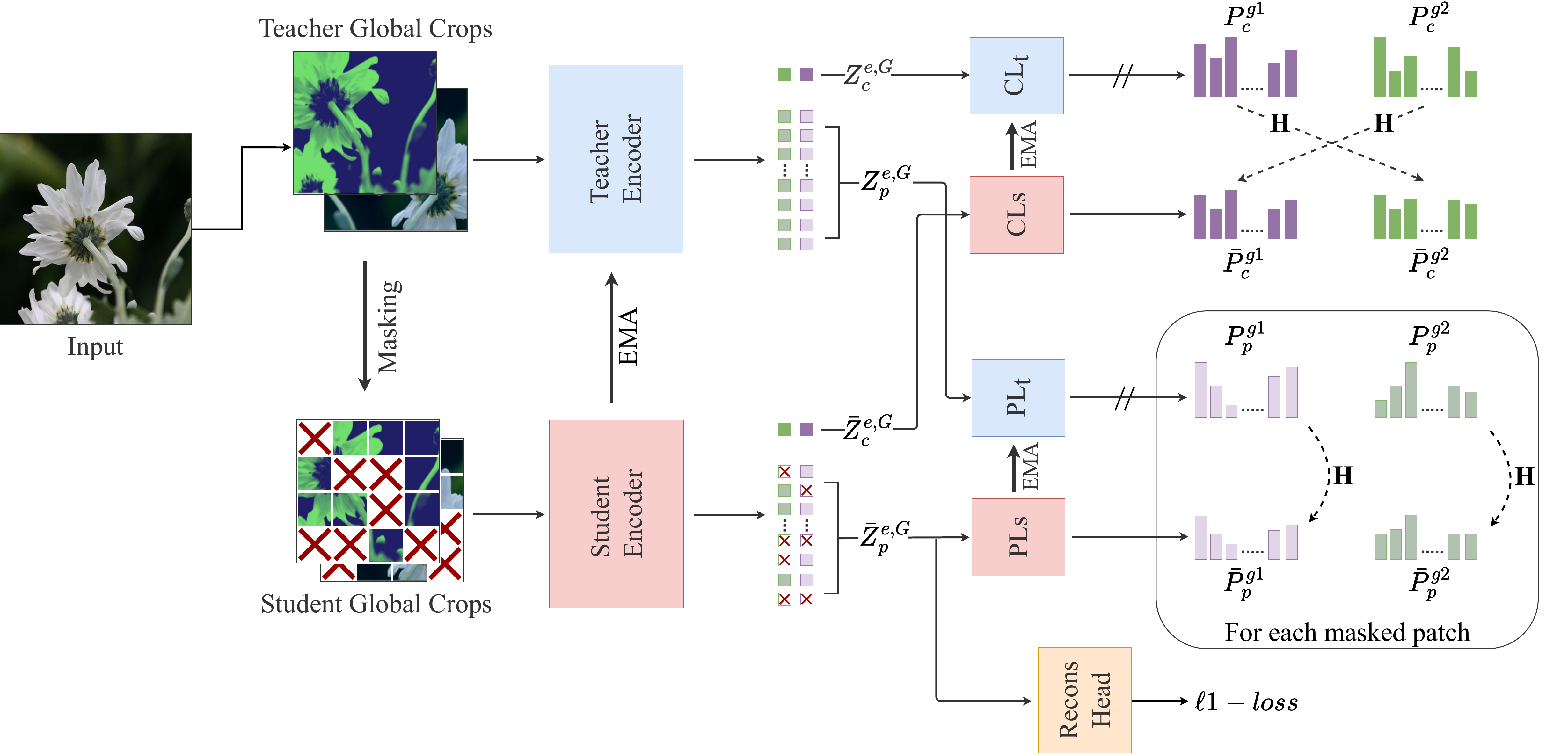}
\caption{
        The MaskCluster architecture for low-shot learning generates multiple masked global views. A teacher encoder, employing transformer layers, produces embeddings $\mathbf{Z}^{e, G}_{c}$ (class tokens) and $\mathbf{Z}^{e, G}_{p}$ (patch tokens) from unmasked global views. Teacher clustering layers $CL_t$ and $PL_t$ assign clusters $\mathbf{P}^{G}_{c}$ and $\mathbf{P}^{G}_{p}$ based on these embeddings. The student encoder similarly processes masked global crops to produce embeddings $\mathbf{\bar{Z}}^{e, G}_{c}$ and $\mathbf{\bar{Z}}^{e, G}_{p}$, which are clustered by student layers $CL_s$ and $PL_s$ to generate assignments $\mathbf{\bar{P}}^{G}_{c}$ and $\mathbf{\bar{P}}^{G}_{p}$. Additionally, $\mathbf{\bar{Z}}^{e, G}_{p}$ undergoes reconstruction to the pixel space using a student's reconstruction head. Losses include cross-entropy between teacher and student cluster assignments, and $\ell$1-loss for reconstructing the original view from the image reconstruction.
        }
\label{fig:arch_overview}
\end{figure*}

\section{Details of SSL components and Analysis}
\label{sec:method}
Here, we provide a detailed analysis of the effect of the choice of pretext tasks, and the choice of a collapse avoidance mechanism on low-shot downstream tasks. In addition we propose an architecture, which is based on the findings of the study. We believe that the choice of pretext task produces a huge impact when finetuning on low-shot tasks. Generally, clustering/contrastive learning only focuses on instance level discrimination and can be classified as an instance discrimination tasks. Focusing on a single global instance might not be beneficial in low data regimes. We believe that finegrained contextual information is necessary for the model to perform better in low data regimes. Collapse avoidance also plays a crucial role when evaluating the self supervised model on low-shot data~\cite{msn}. We therefore present a study on the effects on different collapse avoidance mechanisms.
 
In addition we also answer the question of application of instance discrimination at class token level or to both class and patch level. iBoT~\cite{ibot} shows that application of instance discrimination at both and class level helps for finetuning of full scale data. We avoid the study of different architectures and stick to vision transformer particulary ViT-S~\cite{vit} for fair comparision. Table~\ref{table:comp} presents an overview of different components used by previous SSL frameworks. Based on the detailed study we introduce a low-shot capable self supervised model which also scales to large scale datasets. 

\subsection{Introduction of different SSL pretext tasks:}
We provide a brief introduction to different pretext tasks and their formulation. In this study we focus on contrastive learning, clusering and masked image modelling with pixel reconstruction as the main pretext tasks. Instance discrimination pretext tasks like clustering and contrastive learning are better at learning semantics. MIM based pretext task which reconstruct masked image at pixel level learn local context.  We define theese tasks in the context of vision transformer~\cite{vit}. 

\noindent
\textbf{Contrastive Learning:} Contrastive learning introduced for self supervision in SimCLR~\cite{simclr} generally has $2N$ augmented data points generated from $N$ original images where each image generates two random augmented views. Let $\mathbf{z}_i$ be the output embeddings generated from $i_{th}$ data point after passing through a vision transformer with projection head attached~\cite{simclr}. If $\mathbf{z}_i$, $\mathbf{z}_j$ are the embeddings of positive pairs then the contrastive loss is provided by equation~\ref{eq:lc}.

\begin{equation}
    \mathcal{L}_{con}(i, j) = -log\frac{exp(sim(z_{i}, z_{j})/\tau)}{\sum^{2N}_{k=1}\mathbbm{1}_{[k\neq i]}exp(sim(z_{i}, z_{j})/\tau)}
    \label{eq:lc}
\end{equation}

Here loss $\ell_{con}(i, j)$ is for a single positive pair $(i, j)$ and it will be applied to all the positive pairs including $(j, i)$. $\mathbbm{1}_{[k\neq i]} \in \{0, 1\}$ 
is a function that indicates if $k\ne1$.

\noindent
\textbf{Clustering:} Clustering is a negative sample free pretext task which makes it less sensitive to the choice and number of negative samples within the batch. It also does not require large batch sizes or memory bank used in contrastive learning~\cite{simclr, mocov3}. We generate two random global augmented views $\mathbf{x}_{g1}$, $\mathbf{x}_{g2}$ from an input image $\mathbf{x}$. \citep{dino, ibot} also use several local crops in addition to global crops but we skip them for the loss formulation for simplicity. Generally clustering requires a teacher and student where the teacher is an exponential moving average(EMA) of the student. Let $\mathbf{c}_{g1}$, $\mathbf{c}_{g2}$ be the predicted cluster assignments corresponding to class token of the student after passing through the projection head similar to DINO~\cite{dino}. Let $\mathbf{\bar{c}}_{g1}$, $\mathbf{\bar{c}}_{g2}$ be the target cluster assignments corresponding to class token of teacher after passing through its projection head.  Then the class level clustering loss is defined as cross entropy between $\mathbf{c}_{g1}$, $\mathbf{\bar{c}}_{g2}$ and between $\mathbf{c}_{g2}$, $\mathbf{\bar{c}}_{g1}$.  given by equation~\ref{eq:cls}.

\begin{equation} \label{eq:cls}
    \mathcal{L}_{cc} = \frac{1}{2}(\text{H}(\mathbf{c}_{g1}, \mathbf{\bar{c}}_{g2}) + \text{H}(\mathbf{c}_{g2}, \mathbf{\bar{c}}_{g1}))
\end{equation}

iBoT \cite{ibot} considers extending the clustering loss to patches also for which we provide the formulation in equation \ref{eq:ptch}. Let $\mathbf{p}_{g1}^{i}$, $\mathbf{p}_{g2}^{i}$ be the predicted cluster assignments corresponding to patch token $i\in \{1 \dots N\}$ of student encoder after projection head where N is the number of tokens. Similarly teacher encoder generates target patch cluster assignments  $\mathbf{\bar{p}}_{g1}^{i}$, $\mathbf{\bar{p}}_{g2}^{i}$. 
 
\begin{equation} \label{eq:ptch}
    \mathcal{L}_{pt} = \frac{1}{2N}\sum_{i}^{N}(\text{H}(\mathbf{p}_{g1}^i, \mathbf{\bar{p}}_{g1}^i) + \text{H}(\mathbf{p}_{g2}^i, \mathbf{\bar{p}}_{g2}^i))
\end{equation}

\noindent
\textbf{Mask Image Modelling:}
\newline
Masked image modelling has been explored in SiT \cite{sit} for pixel level reconstruction which has been utilised by several following works~\cite{mae, simmim}. Pixel level reconstruction captures finegrained information required for low data regime~\cite{gmml}. If $x$ is an input image we generate a masked image $\mathbf{x}_m$ with mask $\mathbf{M}$ where $\mathbf{M}(i, j)=1$ if the $\mathbf{x}=(i, j)$ is masked. The masked image $\mathbf{x}_m$ is passed through a vision transformer encoder and a light weight reconstruction head~\cite{simmim} that produces a reconstructed image $\mathbf{\hat{x}}$. The loss for reconstruction is $\ell$1-loss given in equation~(\ref{eq:mim}).

\begin{equation} \label{eq:mim}
    \mathcal{L}_{mim} = \sum_{i}^{H} \sum_{j}^{W} \mathbf{M}(i, j) \times |\mathbf{x}(i, j) - \mathbf{\hat{x}}(i, j)|)
\end{equation}
 
\subsection{Collapse Avoidance:}
Collapse avoidance has been majorly applied to clustering methods like~\cite{dino, msn, ibot}. We mainly study three different methods namely centring explored in DINO~\cite{dino} and iBoT~\cite{ibot}, sinkhorn explored in SwaV~\cite{swav}, and ME-MAX loss introduced in MSN~\cite{msn}. We skip the details of these collapse avoidance to the above mentioned literature but provide a formulation of ME-MAX since we also extend it to patches. MSN studies the effect of ME-MAX only on class token and by default combines it with sinkhorn. ME-MAX is also different from other methods where it is applied as a loss on student cluster assignments whereas both sinkhorn and centring can be considered as a normalisation applied to the teacher target cluster assignments. Let $\mathbf{\bar{c}}$ be the average cluster assignments corresponding to class token,  $\mathbf{\bar{p}}$ be the patch level cluster assignments both generated from the student, we define the ME-MAX at class token level as $\mathcal{L}_{mc} = - \mathbf{H}(\mathbf{\bar{c}})$ and for patch level as $\mathcal{L}_{mp} = - \mathbf{H}(\mathbf{\bar{p}})$ where $\mathbf{H}$ is the entropy function.

\begin{table}[h!]
        \centering
        \resizebox{0.9\linewidth}{!} {
        \begin{tabular}{c c c c c c c } 
        \hline
        Contrastive &  Clustering & MIM & 1-Shot & 2-Shot & 5-Shot\\  
        \hline
        \cmark & \xmark & \xmark & 2.3 & 4.9 & 17.4\\
        \xmark & \cmark & \xmark & 6.2 & 12.9 & 29.6\\
        \xmark & \xmark & \cmark & 8.7 & 17.9 & 32.1\\
        \cmark & \xmark & \cmark & 6.4 &17.0& 30.2\\
        \xmark & \cmark & \cmark & \textbf{15.7} & \textbf{25.6} & \textbf{39.9} \\
        \hline
    \end{tabular}
    }
    \caption{Evaluation of pretext task on ImageNet-$1$K low-shot multi-class classification performance. All the models are ViT-S pretrained with different pretext tasks for $400$ epochs.} 
    \label{table:an_pretext}
\end{table}

\begin{table}[h!]
        \centering
        \resizebox{0.9\linewidth}{!} {
        \begin{tabular}{c c c c c c c } 
        \hline
        ME-MAX &  Sinkhorn & Centring & 1-Shot & 2-Shot & 5-Shot\\  
        \hline
         \xmark & \xmark & \cmark & 5.3 & 12.6 & 28.8\\
         \xmark & \cmark & \xmark & 1.2 & 2.5 & 6.5\\
         \cmark & \xmark & \xmark & 6.2 & 12.9 & 29.6\\
        \hline
    \end{tabular}
    }
    \caption{Evaluation of collapse avoidance mechanisms on ImageNet-$1$K low-shot multi-class classification performance. All the models are ViT-S pretrained with either ME-MAX, sinkhorn or centring for $400$ epochs.}
    \label{table:an_collpase}
\end{table}

\begin{table}[h!]
    \small
        \centering
        \begin{tabular}{c c c c c c c } 
        \hline
        Class & Patch & 1-Shot & 2-Shot & 5-Shot\\  
        \hline
        \cmark & \xmark & 6.2 & 12.9 & 29.6\\
        \cmark & \cmark & 7.9 & 16.5 & 35.0\\
        \hline
    \end{tabular}
    \caption{Evaluation of class level or both class and patch level clustering on INet-$1$K low-shot multi-class classification performance. All models are ViT-S pretrained for $400$ epochs.}
    \label{table:an_cls}
\end{table}

\vspace{-0.3cm}
\subsection{Analysis}
We use ViT-S as our base architecture to study choice of pretext tasks, collapse avoidance on low-shot downstream tasks. We use low-shot classification on ImageNet-1K to study the effect of these choices have on the performance. For 1-shot, 2-shot and 5-shot ImageNet-1K classification we utilise a standard dataset made available from MSN \cite{msn} where each of them have three different splits. All the experiments are pretrained for $400$ and finetuned on target dataset with the accuracy reported for mean of three splits.

\noindent
\textbf{Which pretext task to choose?} We explore clustering and contrastive learning as instance discrimination pretext tasks, along with MIM for pixel reconstruction, emphasizing context. Comparative experiments, detailed in Table~\ref{table:an_pretext}, reveal that clustering with ME-MAX loss outperforms contrastive learning. Focusing on a single semantic object is suboptimal for low-shot performance, as evidenced by inferior results in instance discrimination methods compared to masked image modeling. Clustering, unaffected by the choice of negative samples, surpasses contrastive learning. Combining clustering with MIM yields the best performance, underscoring the importance of fine-grained context and discriminative information for low-shot scenarios.

\noindent
\textbf{Collapse avoidance makes a difference?}
Collapse avoidance is done either through centring~\cite{dino}, ME-MAX~\cite{msn} or sinkhorn~\cite{swav}. We evaluate the effects of all the above methods for collapse avoidance on low-shot evaluation performance in Table~\ref{table:an_collpase}. We find that applying ME-MAX is better compared to sinkhorn and centring. Forcing the network to learn to use all the available clusters at the output through loss is helping the network in low-shot regime.

\noindent
\textbf{Instance discrimination at patch level is needed?}
We assess the impact of patch-level instance discrimination on low-shot performance. Inspired by iBoT~\cite{ibot}, which demonstrates enhanced ImageNet-$1$K finetuning with patch clustering, we explore the potential benefits for low-shot scenarios. In Table~\ref{table:an_cls}, we analyze clustering applied solely at the class level and at both class and patch levels. Given clustering's superiority over contrastive learning (Table~\ref{table:an_pretext}) and the effectiveness of clustering with ME-MAX (Table~\ref{table:an_collpase}), applying clustering at both levels proves more effective for low-shot evaluation than at the class level alone. This emphasizes the significance of local discriminative information, enhancing network performance in downstream tasks.

\vspace{-0.35cm}
\subsection{Simple pretext combination for low-shot}
Based on the above analysis we find that the capturing information at various levels is required for low-shot learning.  Thus we introduce MaskCluster model which captures global and local semantics with clustering pretext task while also having local contextual information from MIM pretext task with pixel level reconstruction. The model has both student and teacher networks both based on vision transformer~\cite{vit}. We attach a projection head similar to previous approaches~\cite{sit, ibot, dino, msn} to generate cluster assignments. The architecture overiew is present in Figure~\ref{fig:arch_overview}. The design of pretext tasks is discussed in previous subsections and our MIM pretext task closely follows GMML~\cite{gmml}. In addition our clustering is done through ME-MAX loss which shows a slight improvement in low-shot setting. Our total loss is provided in equation~\ref{eq:tot} which is the summation of clustering loss at patch and class level, ME-MAX loss on patch and class cluster assignments and finally the MIM reconstruction loss. 

\begin{equation} \label{eq:tot}
    \mathcal{L}_{total} = \mathcal{L}_{cc} + \mathcal{L}_{pt} + \mathcal{L}_{mim} + \mathcal{L}_{mc} + \mathcal{L}_{mp}
\end{equation}

\section{Experimental Results} \label{sec:exp}
We thoroughly evaluate the pretrained model's low-shot performance across three different downstream tasks: multi-class classification, multi-label classification and semantic segmentation. Our approach is further evaluated using the common finetuning evaluation protocol with full datasets. We evaluate by finetuning on diverse downstream tasks, including multi-class classification, multi-label classification, and semantic segmentation, as shown in Section~\ref{sec:train}.



\subsection{Pre-Training Setup}

We use ViT-S/16 as the base backbone architecture pretrained on ImageNet-1K. 
Our approach follows the conventional multi-crop method, with two global crops of resolution $224 \times 224$, and ten local crops of resoultion  $96 \times 96$.
The class, patch clustering layers have a dimension of $8192$. We pretrain the model for $800$ epochs with a $50\%$ masking ratio, following GMML masking strategy~\cite{gmml}. We use the AdamW optimiser \cite{adamw} with weight decay of $0.05$, learning rate of $5e^{-4}$, gradient clipping threshold of 3.0, and 15 warmup epochs.

\begin{table}[h!]
    \centering
    \resizebox{0.99\linewidth}{!}{
        \begin{tabular}{l c c c c c}
            \hline
            Method & Arch. & 1Img & 2Imgs & 5Imgs & 1\% INet\\  
            \hline
            \textit{Rand int.} & ViT-S & 0.7 & 1.5 & 3.8 & 8.5 \\
            SiT \cite{sit} & ViT-S &  6.4 & 17.0 & 30.2 & 51.0\\
            MoCov3 \cite{mocov3} & ViT-S & 2.3 & 4.9 & 17.4 & 50.1 \\
            Dino \cite{dino} & ViT-S & 5.3 & 12.6 & 28.8 & 46.8  \\
            iBoT \cite{ibot} & ViT-S & 7.9 & 16.5 & 35.0 & 50.4 \\
            MSN \cite{msn} & ViT-S & 6.2 & 12.9 & 29.6 & 50.7  \\
            MaskCluster (Ours) & ViT-S & \cellcolor{blue!20}\textbf{22.4} & \cellcolor{blue!20}\textbf{34.8} & \cellcolor{blue!20}\textbf{48.9} & \cellcolor{blue!20}\textbf{58.5}\\ 
            \hline
            \textcolor{gray}{MAE~\cite{mae}} & \textcolor{gray}{ViT-B} & \textcolor{gray}{5.7} & \textcolor{gray}{14.8} & \textcolor{gray}{36.1} & \textcolor{gray}{51.1} \\ 
            \hline
        \end{tabular}
    }
    \caption{The low-shot performance evaluation on subset of INet-1K. All the models are pretrained on INet-1K dataset. We report mean top-1 accuracy of three different splits.}
    \label{table:low-shot}
    \vspace{-0.4cm}
\end{table}

\begin{table*}[t]
\centering
\small
\begin{tabular}{l c c c c c | c | c c c c | c} 
 \hline
 \multirow{2}{*}{Method} & \multirow{2}{*}{Arch.}  & \multicolumn{5}{c|}{1 Shot}  & \multicolumn{5}{c}{5 Shot} \\ \cline{3-12}
  & & Split 0 & Split 1 & Split 2 & Split 3 & Mean & Split 0 & Split 1 & Split 2 & Split 3 & Mean\\ [0.5ex] 
 \hline
  & & \multicolumn{10}{c}{few-shot segmentation methods} \\ \cline{3-12}
 RPMM \cite{rpmm} & Res-50 & 55.2 & 66.9 & 52.6 & 50.7 & 56.3 & 56.3 & 67.3 & 54.5 & 51.0 & 57.3\\
 PFENet \cite{pfenet} & Res-50 & 61.7 & 69.5 & 55.4 & 56.3 & 60.8 & 63.1 & 70.7 & 55.8 & 57.9 & 61.9 \\
 CyCTR \cite{cyctr} & Res-50 & 67.8 & 72.8 & 58.0 & 58.0 & 64.2 & \textbf{71.1} & 73.2 & 60.5 & 57.5 & 65.6 \\
 HSNet \cite{hsnet} & Res-50 & 64.3 & 70.7 & 60.3 & 60.5 & 64.0 & 70.3 & 73.2 & 67.4 & 67.1 & 69.5\\
 BAM \cite{bam} & Res-50 & \textbf{69.0} & \textbf{73.6} & \textbf{67.6} & \textbf{61.1} & \textbf{67.8} & 70.6 & \textbf{75.1} & \textbf{70.8} & \textbf{67.2} & \textbf{70.9} \\
 FPTrans(ViT init.) & ViT-S/16 & 59.7 & 64.0 & 58.5 & 49.2 & 57.8 & 67.8 & 70.5 & 70.2 & 62.4 & 67.7\\ \cline{3-12}
  & & \multicolumn{10}{c}{Self supervised init.(FPTrans)} \\ \cline{3-12}
 Dino \cite{dino} & ViT-S/16 & 39.4 & 40.4 & 30.3 & 31.8 & 35.5 & 49.8 & 57.7 & 49.8 & 44.8 & 50.5 \\
 iBoT  \cite{ibot} & ViT-S/16 & 28.6 & 34.7 & 27.6 & 25.7 & 29.2 & 43.5 & 51.9 & 33.7 & 41.6 & 42.7\\
 MSN  \cite{msn} & ViT-S/16 & 50.3 & 58.5 & 48.4 & 41.8 & 49.7 & 61.2 & 66.7 & 60.6 & 52.9 & 60.2 \\
 MaskCluster (Ours)  & ViT-S/16 & \cellcolor{blue!20}\textbf{68.8} & \cellcolor{blue!20}\textbf{66.5} & \cellcolor{blue!20}\textbf{65.4} & \cellcolor{blue!20}\textbf{53.2} & \cellcolor{blue!20}\textbf{63.5}& \cellcolor{blue!20}\textbf{73.6} & \cellcolor{blue!20}\textbf{71.3} & \cellcolor{blue!20}\textbf{72.3} & \cellcolor{blue!20}\textbf{62.8} & \cellcolor{blue!20}\textbf{70.0}\\ 
 \hline
\end{tabular}
\caption{The Pascal 5i low-shot segmentation results. We compare few-shot segmentation methods and self supervised methods including ours that are used to initialise the ViT-S/16 of FPTrans. FPTrans(ViT init.) represents ViT-S/16, initialised with Imagnet-1K supervised weights.}
\label{table:fewseg}
\end{table*}

\begin{table}[h!]

\centering
\small
\begin{tabular}{l c c c | c}
\hline
\multirow{2}{*}{Method}  & \multicolumn{3}{c|}{Pascal VOC} & \multirow{ 2}{*}{Mini.COCO}\\ \cline{2-4}
  & 1Img & 2Imgs & 5Imgs\\ \hline
 MoCoV3 \cite{mocov3} & 18.6 & 19.3 & 22.2 & 50.7 \\
 DINO \cite{ibot} & 23.9 & 34.2  & 42.6 & 50.1 \\
 iBoT \cite{ibot} & 25.2 & 35.4 & 44.0 & 50.4 \\
 MSN \cite{msn} & 23.0 & 30.8 & 40.0 & 50.9 \\
Ours  & \cellcolor{blue!20}\textbf{27.1} & \cellcolor{blue!20}\textbf{40.7} & \cellcolor{blue!20}\textbf{46.6} & \cellcolor{blue!20}\textbf{54.1}\\ 
 \hline
\end{tabular}
\caption{Low-shot multi-label classification on Pascal VOC  and mini COCO with mAP as a metric. All the SSL models are pretrained on INet-1K, employing ViT-S/16 backbone.}
\label{table:low-shot_multi}
\end{table}

\subsection{Main Results}\label{sec:train}
\noindent
\textbf{Low-shot Multi-class Classification.}
To assess label efficiency, we fine-tune our model on smaller subsets of ImageNet-1K, utilizing MSN's~\cite{msn} data subsets for 1, 2, or 5 images per label. Each of these datasets have three different splits and we evaluate on all the splits while reporting the mean accuracy. Furthermore, we finetune on 1\% of ImageNet-1K based on the SimCLR~\cite{simclr} split. Results (Figure~\ref{fig:plot_low-shot}) highlight our approach's substantial performance lead over current state-of-the-art techniques, particularly under extremely limited data during fine-tuning. For instance, with only 1 image per label, our method attains 22.4\% accuracy, surpassing the state-of-the-art by 14.4\%. Suprisingly we also perform better when compared to MAE~\cite{mae} with much larger ViT-B as the encoder. The results are provided in Table~\ref{table:low-shot}. 

\begin{table}[h!]
\centering
\resizebox{0.9\linewidth}{!} {
\begin{tabular}{l  c c c c c | c} 
 \hline
 Method  & Flwrs & Pets & Cars & C-10 & C-100 & INet-1K\\ 
 \hline 
 Supervised & 98.2 & -- & 92.1 & 99.0 & 89.5 & 79.9\\
MoCov3 \cite{mocov3} & 97.7 & 92.3 & 93.0 & 98.2 & 86.6 & 81.4\\ 
DINO \cite{dino} & 97.8 & 89.4 & 93.1 & 99.0 & 90.5 & 81.5\\
iBoT \cite{ibot} & 98.6 & 93.1 & \textbf{94.0} & 99.1 & 90.7 & \textbf{82.3} \\
Ours  & \cellcolor{blue!20}\textbf{98.7} & \cellcolor{blue!20}\textbf{93.7} & \cellcolor{blue!20}93.6 & \cellcolor{blue!20}\textbf{99.2} & \cellcolor{blue!20}\textbf{90.8} & \cellcolor{blue!20}82.1 \\ 
 \hline
\end{tabular}
}
\caption{Transfer learning by finetuning pretrained models with the ViT-S/16 backbone on diverse datasets. We report top-1 accuracy.}
\label{table:small_finetune}
\end{table}

\noindent
\textbf{Low-shot Semantic Segmentation.}
To assess low-shot segmentation performance, we employ FPTrans \cite{fptrans}, a framework tailored for few-shot segmentation with vision transformers. Evaluation is conducted on the Pascal 5i dataset under two scenarios: one-shot and five-shot. Leveraging pre-trained SSL method weights to initialize ViT-S/16 within FPTrans, training settings align closely with those specified in FPTrans. Additionally, we train FPTrans using supervised ImageNet-1K weights for ViT-S/16, enabling comparative analysis. Our model not only surpasses other self-supervised methods but also demonstrates competitive performance in low-shot segmentation across both one-shot and five-shot settings (see Table \ref{table:fewseg}). These findings highlight our model's outstanding capability in low-shot semantic segmentation.

\noindent
\textbf{Low-shot multi-label Classification.}
Datasets for 1, 2, and 5 images per label, along with Mini.COCO~\cite{minicoco} dataset (20\% of MS-COCO training data), are created by random sampling. Methods are trained with an input resolution of $224 \times 224$, and comparisons in Table~\ref{table:low-shot_multi} show our model's superiority in multi-label classification efficiency over SSL methods.

\noindent
\textbf{Multi-class Classification.}
We evaluate the performance of our method by finetuning pretrained weights from ImageNet-1K on various downstream datasets (Table \ref{table:small_finetune}). Our model achieves $82.1\%$ top-1 accuracy on INet-1K, which is comparable to state-of-the-art methods. Additionally, we showcase the effectiveness of our approach through finetuning results on smaller datasets like Flowers, Cars, Pets, CIFAR-10, and CIFAR-100. Across all these datasets, our method consistently outperforms state-of-the-art approaches, demonstrating the strong transferability of our proposed framework.

\vspace{-1em}
\section{Conclusion}\label{sec:conclusion}
In this study, we investigate the influence of different pretext tasks and strategies to avoid collapse on low-shot performance. Clustering outperforms contrastive learning, especially for instance discrimination tasks. Combining reconstruction with instance discrimination, particularly through clustering, enhances low-shot performance. Further improvements are observed when applying clustering at both class and patch levels. Drawing from these findings, we propose a multi-level architecture leveraging clustering for global and local semantics, alongside reconstructing masked images. This architecture excels in low-shot downstream tasks and scales effectively to full dataset fine-tuning across multiple tasks. Future work aims to extend the low-shot performance of this model to multi-modal settings.

\noindent
\textbf{Acknowledgements:}
This work was supported in part by the EPSRC grants MVSE (EP/V002856/1) and JADE2 (EP/T022205/1).


{
\bibliographystyle{plainnat}
\bibliography{main}
\small
}

\end{document}